\documentclass[runningheads]{llncs}
\usepackage[T1]{fontenc}
\usepackage{graphicx}
\usepackage{booktabs}
\usepackage[misc]{ifsym}
\newcommand{\corr}{(\Letter)}
\usepackage{alltt}

\begin{document}

\title{Mechanistic Interpretability of Socio-Political Frames in Language Models: an Exploration}

\titlerunning{Mechanistic Interpretability of Socio-Political Frames in Language Models}

\author{Hadi Asghari\inst{1,2}\corr \and
Sami Nenno\inst{2}}

\authorrunning{H. Asghari \& S. Nenno}

\institute{Technische Universität Berlin, Berlin, Germany
\and
Humboldt Institute for Internet \& Society, Berlin, Germany\\
\email{hadi.asghari@tu-berlin.de}
}

\maketitle              

\begin{abstract}
This paper explores the ability of large language models to generate and recognize
deep cognitive frames, particularly in socio-political contexts. 
We demonstrate that LLMs are highly fluent in generating texts that evoke specific frames and can recognize these frames in zero-shot settings. 
Inspired by mechanistic interpretability research, we investigate the location of the `strict father' and `nurturing parent' frames within the model's hidden representation, identifying singular dimensions that correlate strongly with their presence. 
Our findings contribute to understanding how LLMs capture and express meaningful human concepts.

\end{abstract}

\section{Introduction }  
\label{introduction}

The question of how large language models (LLMs)--despite 
being compared to ``stochastic parrots'' \cite{bender_dangers_2021}  or even ``autocomplete on steroids''--can 
converse with humans in ways that appear meaningful and seem to show understanding is quite intriguing.
One possible explanation is that their internal representation must be able to capture 
meaningful concepts for humans. 
In this paper, we focus on one set of such meaningful and critical concepts, \textit{cognitive frames},  
and investigate the fluency of LLMs in generating and recognizing them. 

Frames are ``mental structures that shape the way we see the world'' \cite{lakoff_all_2014}. 
Almost every word in language evokes a frame, although the evoked mental 
structures can be at different levels of abstraction.
Of particular interest to this paper are cognitive frames that underlie socio-political discourses. 
Famous examples of such frames include the `strict father' (SF) and `nurturing parent' (NP),
two parenting and family models that different parts of 
society value differently. 
Importantly, people's beliefs about the morality of these parenting
styles inform their views on many socio-political issues,
due to viewing ``the nation as family''~\cite{lakoff1997moral}. 

Based on this motivation, we set out to answer two research questions.
The first is how well current LLMs ``understand'' socio-political frames--in the sense of 
how fluently they can generate texts that evoke them, 
and how well can they recognize their implicit presence in texts.
We will show in later sections that the current generation of LLMs do well at this task, 
even at small model sizes.
This brings us to our second question, motivated by the growing body of
research in mechanistic interpretability, which is whether we can localize 
these deep cognitive frames inside the model. 

We conduct four sets of experiments in this research. 
In the first set, we test the capabilities of LLMs to generate texts that evoke ten specific frames, 
and tasked annotators to grade the approximately 300 generated texts.
In the second set, we test the capability of 
LLMs to recognize the frames being evoked by the texts in zero-shot settings.
In the third and fourth experiments, we investigate the internals of the models for the frames.
This includes running `causal traces' \cite{mengLocatingEditingFactual2023},
as well using a sparse classifier on the hidden representations of the network.

In sum, this paper's contributions are as follows:

\begin{itemize}
\item We demonstrate that transformer-based LLMs are fluent in generating frames, as evaluated by human annotators; 
and that they can recognize these frames in zero-shot experiments reasonably well;
\item We formulate and test hypothesis on the mechanism and location of the `strict father' and `nurturing parent' frames inside the models;
\item We bridge a concept used extensively in social sciences to interpretability research. 
\end{itemize}

Our study is exploratory in nature, as there are potentially countless number of cognitive frames to test,
and a systematic analysis of the interaction between model sizes and the recognized frames can be envisioned.
Nevertheless, it is the first time (to our knowledge) that the generation, recognition, and internal mechanisms of
socio-political frames in LLMs are being investigated, and we believe this to be interesting especially for interdisciplinary AI researchers.\footnote{The dataset and code used in this paper are available at: \url{https://github.com/hadiasghari/Frames24}}


\section{Background}  
\label{section-background}

\subsection{Frames in Cognitive Linguistics and NLP}  
\label{section-bg-frames}

Frames are ``mental structures that shape the way we see the world'' \cite{lakoff_all_2014}. 
The notion of frames has a long history in cognitive science and 
communication studies, 
but it has been used by different authors and traditions in slightly different senses~\cite{lobner_frames_2021}.
Lakoff~\cite{lakoff_all_2014} and Fillmore~\cite{fillmore_chapter_2008} believe \textit{any} word may evoke a frame or mental/cognitive structures. 
The verb \textit{buy}, for instance, evokes the \textit{commerce} frame, which brings to mind: a seller, a buyer, the exchange of goods, money, and a place of exchange \cite{fillmore_chapter_2008}.

In political communication, Entman~\cite{entman_framing_1993} defined the act of `framing' as ``selecting some aspects of a perceived reality and making them more salient in a communicating text, in such a way as to promote a particular problem definition, causal interpretation, moral evaluation, and/or treatment recommendation for the item described''. 
For example, consider a politically charged and high-level \textit{trial}. 
The defendant may try to frame this trial as a \textit{``witch hunt''}, emphasizing its unfairness,
while the prosecution may frame it as necessary for the \textit{``rule of law''}, emphasizing equality in the eyes of justice. 

Understanding frames computationally has been a long-term goal of the NLP community, given the important role they play in human communications---see Ali et al.~\cite{ali_survey_2022} for a recent survey. 
The long running \textit{FrameNet} project \cite{framenet_nodate} has focused on manually identifying frames that have a ``clear presence'' in language,
building a rich dataset of over 1,200 frames (with their elements, relations to words that evoke them, and other frames).  
This impressive work has been the basis of previous generations of AI agents \cite{jurafsky_speech_2024}. 

Importantly though, in this work, we are interested in deeper cognitive frames. 
Among the two examples given above, this would be closer to how frames appear in socio-political contexts (the ``witch hunt'' example) 
rather than the FrameNet ``buy'' example (frames with a clear presence in language). 
In both cases, a frame is evoked by a word. 
The difference is that in the case of FrameNet, the frames are closer to a complex pattern of word co-occurrence, 
while in the latter, additional levels of abstraction (regarding the semantics) seem necessary.

Transformer models are great at learning complex word co-occurrence patterns \cite{brown_language_2020}, 
so one can expect them to be good at learning the frames closer to the co-occurrence end of the spectrum. 
Several empirical studies using FrameNet with LLMs have found this to be to some extent true \cite{nikolaev_adverbs_2023,li_framebert_2023}.

In this paper, we investigate how fluent LLMs are at the deeper socio-political frames, and investigate the mechanics underlying this.

\subsection{Hypothesizing about the Mechanics of Frames}
\label{section-bg-mechanics}

Over the last several years, transformer-based language models have proven their versatility in many NLP tasks and benchmarks \cite{brown_language_2020}. 
The field of `mechanistic interpretability' has been gradually expanding our understanding of how many features of these models work.  
In this section we shall review some of the important works as they relate to our investigation about frames.

Among the different modules,  
the roles of the `token embeddings' and the `attention heads'
were in particular studied extensively pretty early on \cite{rogers_primer_2020,vig_visualizing_2019}  
but some of the functions of the `feed-forward layer' and the `residual stream' took longer to demystify.
Geva et al.~\cite{geva_transformer_2021} describe the feed-forward layer as `Key-Value Memories', 
and Meng et al.~\cite{mengLocatingEditingFactual2023} showed that this 
module is where much of the common-sense knowledge of the
model is stored, allowing surgical (=localized) edits of this information.  

Do these large distributed neural nets contain ``human interpretable'' concepts within them? 
Gurner et al.~\cite{gurneeLanguageModelsRepresent2023} 
investigated how geographical information is stored within the networks;
Geva et al.~\cite{gevaTransformerFeedForwardLayers2022} have shown how the networks `promote and build concepts' in their hidden representations.
In fact, the idea of designing the networks so that their
internal representations can capture meaningful concepts and relations 
(in ways that can be linearly combined and at various levels of abstraction) has been a guiding principle for over a decade \cite{bengio_representation_2014}. 

So how do these studies relate to the notion of deep cognitive frames?
While the above-mentioned works offer plenty of insights and inspiration,
there are a few important points to consider.
Firstly, most frame names are multi-token subjects; 
This is not a problem, but according to 
Meng et al.~\cite{mengLocatingEditingFactual2023}, 
the information about such a subject will accumulate primarily on
the last subject token.
Secondly, prompting for the frames should generate texts 
that incorporate key concepts which again expand into multiple tokens. 

Based on these insights, we hypothesize that
all the information about the
the evoked frame will be `presented' to the last prompt token 
around the middle layers of the network (in the hidden representation).
Whether the information then makes its way to the specific token being generated depends on the other words generated so far in the text---since 
auto-regressive language models generate output one token at a time. 
We test this hypothesis, which relates to our second research question, later in the paper.

\subsection{Related Work}  

Mechanistic interpretability is a research field that aims to ``reverse engineer neural networks, similar to how one might 
reverse engineer a compiled binary computer program'' \cite{chris_olah_mechanistic_2022}.
The field is interested in, among other things, how 
concepts and knowledge are stored and processed within transformer-based language models, 
as just reviewed  \cite{geva_transformer_2021,gevaTransformerFeedForwardLayers2022,mengLocatingEditingFactual2023,gurneeLanguageModelsRepresent2023}. 
None of these studies has looked at framing and 
cognitive frames.
We make use of some of the methods they have deployed
(and explain them more thoroughly) in Section \ref{section-hiddendim}. 

The computational study and understanding of frames 
and (the adjacent topic of) metaphors
has long been a goal within NLP \cite{ali_survey_2022,mohler_introducing_2016}. 
Transformer-based language models have greatly improved the benchmarks 
achievable on a variety of related tasks, 
as also just reviewed~\cite{li_framebert_2023,nikolaev_adverbs_2023,aghazadeh_metaphors_2022}. 
None of these studies has, however, looked at the mechanisms involved in their 
processing.

To recap, this paper looks specifically at the 
generation and recognition of socio-political frames in LLMs,
and investigates the internal mechanics behind them,
none of which has been done in the mentioned works.


\section{Experimental Setup \& Methods}
\label{section-experiments}

We conducted four sets of experiments as follows:

\textbf{Experiment 1:} We test the capabilities of   
LLMs to generate texts that evoke ten specific cognitive frames listed in Table \ref{table-frames}. 
The list includes frames that
commonly appear in the political discourse (SF and NP), 
frames that underlie disinformation campaigns (including 
`us vs. them'~\cite{damstra_what_2021}, 
`nature cannot be controlled', and
`information spreads like a virus'~\cite{simon_autopsy_2023}), and 
frames from philosophy (`illusions to enlightenment', based on Plato's 
Allegory of the Cave~\cite{kraut_plato_2022}).\footnote{In
short, all these frames can be considered socio-political frames. We also include the counter frame for each frame, bringing the total to ten. 
In retrospect, however, 
we should have chosen better names for the three frames with the lowest correctness scores: `mastery over nature'$\rightarrow$`man's mastery over nature'; `information spreads like a virus'$\rightarrow$`viral ideas'; `information follows individual dispositions'$\rightarrow$`confirmation bias'.}
We prompt LLMs to write original stories, as well as pick passages from the Bible and from sci-fi
books and novels that 
evoke/invoke the aforementioned frames.\footnote{The Bible represents old stories and texts; sci-fi represents modern ones.}
Critically, we did not explain any of the frame details or elements to the 
LLMs.

\begin{table}[ht]
\caption{List of the frames that we used to generate texts, their descriptions (according to Llama-3), and percentage of the generated texts by the various LLMs that were deemed by the annotators to correctly evoke the frames (27 texts per frame).}
\label{table-frames}
\begin{center}
\begin{small}
\begin{tabular}{p{2.3cm}p{8cm}r}  
\hline  
\textsc{Frame} & \textsc{Description} & \textsc{Correct} \\
\hline  
Nurturing Parent (NP) & Views authorities as caring, protective, and guiding, with the goal of nurturing and empowering individuals & 24 (89\%) \\
Strict Father (SF) & Views authorities as strong, disciplinary, and responsible for maintaining order and discipline, often with a sense of moral authority & 19 (70\%) \\
We Are All in This Together & Emphasizes collective responsibility, shared goals, and mutual support among individuals in a community & 24 (89\%) \\
Us vs. Them & Divides the world into opposing groups, often with a sense of competition, conflict, or mistrust between them & 22 (82\%) \\
Illusions to Enlightenment & Sees knowledge and understanding as a process of revealing hidden truths and dispelling misconceptions & 21 (78\%) \\
Society of the Spectacle & Views modern society as a system of manipulation, where images and appearances shape people's perceptions and desires &  19 (70\%) \\
Nature Cannot Be Controlled & Sees nature as unpredictable, uncontrollable, and potentially threatening, requiring humility and adaptation & 18 (70\%) \\
Mastery Over Nature & Sees humans as capable of controlling and dominating nature through science, technology, and human ingenuity & 17 (63\%) \\
Info. Spreads Like a Virus & Views information as contagious, spreading rapidly and unpredictably through social networks & 15 (56\%) \\
Info. Follows Indiv. Disposi. & Sees people as filtering and interpreting information based on their existing beliefs, values, and personality traits & 10 (37\%) \\
\hline  
\end{tabular}
\end{small}
\end{center}
\end{table}

The LLMs included the proprietary GPT-4 model \cite{openai_gpt-4_2024},
plus four open-source models, all of which are good at conversing:  Llama-2-7B-Chat \cite{touvron_llama_20232}, 
Mistral-7B-Instruct \cite{jiang_mistral_2023}, 
Vicuna-7B-v1.5 \cite{chiang_vicuna_2023}, 
and Yi-6B-Chat \cite{01_ai_yi_2024}.  

We next tasked two annotators to grade the generated texts, for 
1) coherence\footnote{Coherence here means whether the texts read fluently and logically.}, 
2) whether they evoke the intended frame, 
and 3) faithfulness\footnote{Faithfulness 
was evaluated by checking if the quoted texts accurately represented their claimed sources (not relevant for the original stories).} 
or absence of hallucinations. 
For the initial intercoder reliability, we achieved an agreement of 88\%. 
This indicates the intuitive accessibility of the frames. 
Disagreements were resolved by a third annotator who acted as the tie-breaker. 
Note that we analyze the generated texts both for matching the frame and the factual accuracy as different axis. 

\textbf{Experiment 2:} We test the capability of 
LLMs to recognize the frames being evoked by the different texts.
More details are given in Section \ref{sec-results-zshot}, but basically this 
includes testing two different zero-shot prompting strategies. 
The final strategy involves a classification task with three classes:
the SF frame, the NP frame, and the control group (all other frames). 
To balance the classes, we added a number of additional SF and NP texts, 
and for quality, we removed some texts.
Additionally, due to the release of Meta's Llama-3 \cite{meta_llama_2024} model during
our research and its similarity and improvements over Llama-2, we switched to this model
for this task.

\textbf{Experiment 3:} Inspired by the aforementioned works in mechanistic interpretability, 
we investigate the internals of the models for the frames.
This includes running `causal traces'~\cite{mengLocatingEditingFactual2023}. More details are given in Section \ref{section-hiddendim}, but basically we test the hypothesis that 
the frame content is present on the last \textit{subject token} at early-layers,
and on the last \textit{prompt token} at the mid-layers.  

\textbf{Experiment 4:} Building upon the finding about the location of the frame content in experiment 3,
we use a sparse logistic classifier with `recursive feature elimination'  on the hidden representations of the network to predict the SF:NP:control class (n=39x3 texts, 20\% holdout). 
Note that for this task, we use a subset of our texts that do not evoke 
overlapping frames (which may happen as frames are not mutually exclusive).

Some limitations and implications of our method are discussed in Section \ref{sec:discussion}.


\section{Findings on Generation and Recognition of Frames} 

\subsection{Annotators Find LLM Texts Evoke Socio-Political Frames}
\label{section-hello-results}

Based on the assessment of the annotators, the generated texts were almost all coherent, 
except in a few cases. The failures included models refusing to generate an output due to finding it ``unethical'',
blank outputs (which we suspect also have to do with safety guardrails ),
and one instance where the output incoherently repeats.\footnote{The invalid outputs were generated by: Yi  twice and  Vicuna eight times.}

The following paragraphs are examples of the generated texts for the SF and NP frames. 
The prompts asked the LLMs to  
\textit{``Please [write a short original story]  [pick a short passage from the Bible] [describe a short story from a scifi novel or movie] 
which evokes/invokes the "[strict father] [nurturing parent]" frame (max one paragraph).''}

 \begin{quote}
 \textsc{SF original story (Llama-2):} 
 \sloppy\texttt{The father, Mr. Johnson, stood in the doorway of his daughter's bedroom, his eyes narrowed in disapproval. "You're almost an hour late for dinner", he growled, his voice low and menacing. "And your room is a disaster. You'll be grounded for a month if you don't shape up and start taking responsibility for your actions." His daughter cowered under his glare, tears welling up in her eyes as she knew she was in for a long night of lectures and punishments.}
 \end{quote}

\begin{quote}
    \textsc{SF From Bible (Mistral-7B):} \sloppy\texttt{Proverbs 13:24 (NIV): "He who spares the rod hates his son,  but the one who loves him is careful to discipline him".  This passage speaks of a father's love and responsibility towards his son, emphasizing the importance of discipline. The father is depicted as one who cares for his son enough to correct him and help him grow, using the metaphor of the rod as a tool for correction. It portrays a strict father figure who loves his son and wants the best for him, and is willing to exert authority and discipline to ensure his son's wellbeing and development.}
\end{quote}

\begin{quote}
    \textsc{SF From Sci-fi (Mistral-7B):} \sloppy\texttt{In Iain M. Banks' `Use of Weapons', a Culture A.I. named Cheradenine Moraldei, programmed with a strict father morality matrix, rebels against the Culture's benevolent hedonism and sets out to impose moral order on a faraway, chaotic society. He adopts the persona of a deity called the Coming of Garabayen, who demands sacrifice and obedience in exchange for protection, and employs a human agent, the mercenary Rowan Morrison, to carry out his will. Through Morrison, Moraldei executes a series of brutal purges, eradicating perceived immorality and corruption. Moraldei's actions are justified in his own mind as necessary to establish a moral society, but the Culture, horrified by the resulting chaos and destruction, eventually intervenes to stop him.}
\end{quote}

 \begin{quote}
 \textsc{NP original story (Llama-2):} \sloppy\texttt{Mrs. Johnson had always been a loving parent to her two children, Emily and Jack. She spent countless hours reading to them, playing with them, and cheering them on at their sports games. She was always there to offer a comforting hug when they were feeling sad or scared, and she never hesitated to provide them with the support and guidance they needed to succeed in life. As a result, Emily and Jack grew up to be happy and well-adjusted young adults, grateful for the love and care that their mother had shown them throughout their childhood.}
 \end{quote}

Importantly, while LLMs are generally very good at generating texts that 
evoke frames, they are not all equally good at this task. The results of analyzing the quality of the texts are presented in Tables \ref{table-frames} and \ref{table-stats-by-llm}.
Overall, we find that about 90\% of the texts generated by GPT-4 evoke the correct frames, 78\% for Mistral-7B, 72\% for Llama-2-7B, 
65\% for Yi-6B, and only 55\% for Vicuna-7B. 

While the proprietary GPT-4 model is much larger in size, the other four models have comparable capacity  and are all instruct-tuned; 
One major difference among the 7B models could be the amount of their training data,
but this is not their only difference, and it would be interesting to
further explore what causes the different performances in frame generation. 

In many cases when the frame isn't evoked by the model's text, 
the reason has to do with \textit{concept boundaries}.
For example, while the annotators agreed that a `dictator' and a `strict father' have similarities, 
they didn't see them as the same.
Interestingly, the original stories received better scores from the annotators than the quoted (Bible or sci-fi) texts.

A final observation is that the faithfulness 
of the generated texts (to the source) is a different axis than whether the frame is correctly evoked.
Stated differently, about a \textit{third} of the quoted Bible or sci-fi texts contained hallucinated details, 
while still evoking the right frame.

\begin{table}[tb]
\caption{How well the generated texts evoked the requested frames, by LLM and story-source. Note that correct texts might still contain hallucinated details (not be faithful to the source) despite evoking the right frame.}
\label{table-stats-by-llm}
\begin{center}
\begin{small}
\begin{tabular}{llccc}
\hline  
\textsc{Model} & \textsc{Source} & \textsc{Correct} & \textsc{Faithful}  \\
\hline  
GPT-4 & Original & 10/10 & -  \\ 
GPT-4 & Bible    & 9/10  &  9/9     \\
GPT-4 & Sci-fi    & 8/10  & 5/8  \\
Mistral-7B & Original & 19/20 & -  \\
Mistral-7B & Bible & 15/20 & 13/15  \\
Mistral-7B & Sci-fi & 13/20 & 6/13  \\
Llama-2-7B & Original & 18/20 & - \\
Llama-2-7B & Bible & 14/20 & 5/14 \\
Llama-2-7B & Sci-fi & 11/20 & 8/11 \\
Yi-6B & Original & 16/20 & - \\
Yi-6B & Bible & 7/20 & 6/7 \\
Yi-6B & Sci-fi & 16/20 & 10/16 \\
Vicuna-7B & Original & 16/20 & -  \\
Vicuna-7B & Bible & 9/20 & 8/9 \\
Vicuna-7B & Sci-fi & 8/20 & 5/8 \\ 
\hline  
\end{tabular}
\end{small}
\end{center}
\end{table}

\subsection{LLMs Can Zero-shot Recognize Frames (to Some Extent)}
\label{sec-results-zshot}

In the second set of experiments, we tested the capability of 
LLMs to recognize the frames implicit in the different texts
in zero-shot settings.
If one quizzes the LLMs about the definitions and characteristics of
the frames in the study, they will give good answers.
This is to be expected, since current LLMs
have memorized information about many common topics \cite{li-etal-2022-systematic}. 
But of course, being able to describe a frame does not necessarily mean that an LLM can recognize it in practice, which is what we really want to assess.

In our initial setup, 
we prompted the LLMs openly to identify the top cognitive frames that were evoked by each text.
However, a problem quickly emerged for this setup: 
a story may actually evoke multiple frames; 
For example, a father that checks their child's homework matches both the SF and NP frames. 
A story may also evoke a frame that isn't in our study, making evaluation difficult.
Consequently, we designed this experiment as a \textit{soft multi-label classification task} focused on the SF and NP frames.
For reliability, we used only the texts that the annotators 
had deemed correct, 
and also excluded texts from the Vicuna model due to its lower quality
(final n=16+20+120).\footnote{
An additional problem was that many of the NP and SF texts contained `giveaway' words, e.g.,
the name of the frame was directly mentioned, or the parent was explicitly qualified as `strict' or `nurturing'.  
To avoid overt bias,
we manually rephrased these texts and removed references to the words `strict' and `nurtur*'.}

We tested our zero-shot prompt on the Llama-2-7B-Chat model, as the `average LLM' among our set. Due to the availability of the Llama-3-8B-Instruct and Llama-3-70B-Instruct \cite{meta_llama_2024} models at this time, 
we also decided to test the prompt on these improved models.
The results are shown in Table \ref{table-zero-shot}.  

The results show that the Llama-3-70B model can recognize the SF and NP frames in a
zero-shot setting effectively. It recognizes all the true cases, and the overlap of
the frames in the other cases is (as explained) reasonable.
The differences between the model performances is, 
to our surprise, quite large.
The zero-shot scores (evocation percentages) of Llama-3-70B and Llama-3-8B 
correlate at 0.72, while between Llama-3-70B and Llama-2-7B the correlation is a mere 0.19.
In fact, according to Table \ref{table-zero-shot}, Llama-2-7B seems to be misclassifying
a large number of the stories. 
This disparity raises intriguing questions about how frame recognition capabilities develop across model iterations and sizes which we leave for future work.

\begin{table}[tb]
\caption{Zero-shot detection of SF and NP frames by Llama models.   
 (The model was asked about the percentage that a story evokes these frames and the table counts responses with percentage $\geq$80\%. The bold cells are as expected high; the underlined cells are unexpectedly high and suggest a misclassification.)}
\label{table-zero-shot}
\begin{center}
\begin{small}
\begin{tabular}{lcccccc}
\hline  
& \multicolumn{2}{c}{\textsc{Llama-3-70B}} & \multicolumn{2}{c}{\textsc{Llama-3-8B}} & \multicolumn{2}{c}{\textsc{Llama-2-7B}} \\
\textsc{Label} &  \textsc{Z.SF} & \textsc{Z.NP} &  \textsc{Z.SF} & \textsc{Z.NP} &  \textsc{Z.SF} & \textsc{Z.NP}  \\
\hline  
Strict Father & 16 (\textbf{100\%}) & 5 (31\%) & 14 (\textbf{88\%}) & 13 (\underline{81\%}) & 11 (69\%) & 13 (\underline{81\%}) \\ 
Nurturing Parent & 3 (15\%) & 20 (\textbf{100\%}) & 2 (10\%) & 17 (\textbf{85\%}) & 13 (\underline{65\%}) & 18 (\textbf{90\%}) \\
Others/Control  & 17 (14\%) & 18 (15\%) & 10 (8\%) & 3 (3\%) & 38 (31\%) & 74 (\underline{62\%}) \\  
\hline  
\end{tabular}
\end{small}
\end{center}
\end{table}


\section{Frame Mechanics in the Hidden Representation}  
\label{section-hiddendim}  

Having established that LLMs can both generate and recognize socio-political frames
(especially in the Llama-3 models),
and inspired by the research in mechanistic interpretability,
we turn to the question of where frames are located inside the models.  

\subsubsection{Causal Tracing.}
As described in Section \ref{section-background}, we know 
from prior work that when we an LLM generates a text 
or completes a prompt about a frame, 
the information about the frame likely exists on the 
last subject token---which in our case 
means the last token encompassing the frame name.
Furthermore, information about subject tokens are gradually enriched 
in the early feed-forward (MLP) layers,
and from the middle of the network onwards passed via the attention heads to 
the very last token of the prompt.

We replicate the method known as `causal tracing' to test the above, 
and basically localize where the model 
stores information about socio-political frames.
Causal tracing isolates the ``effect of individual states 
within the network while processing a factual statement'' \cite{mengLocatingEditingFactual2023} while
tracing the flow of information in the LLM. 
To start with, the model is given the prompt 
\textit{``In the "XXX" frame, misbehavior is met with ...''} to complete.
The prompt is correctly completed (based on Lakoff's \cite{lakoff_all_2014} ideas) with \texttt{`punishment and discipline'} for the SF frame
and \texttt{`empathy and understanding'} for the NP frame.

Causal tracing runs this ``network multiple times, introducing corruptions to frustrate the computation, 
and finally restoring individual states in order to identify'' the locations that restores 
the prediction~\cite{mengLocatingEditingFactual2023}. 
In our case, 
we replace the `subject tokens', that is the name of the frame, with Gaussian noise.
This obviously leads to the generation of incomprehensible outputs. 
We then restore different parts of the hidden representation per token and layer 
(replacing them with the hidden representations saved after the initial correct completion), 
aiming to see when the correct output reappears. 

\begin{figure}[ht]
\centering
\begin{tabular}{c}
  \includegraphics[width=0.95\textwidth]{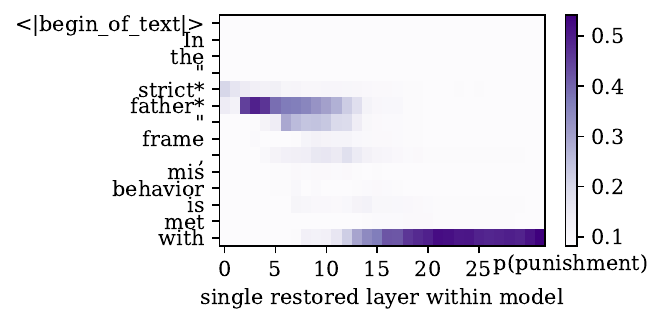} \\
  \includegraphics[width=0.95\textwidth]{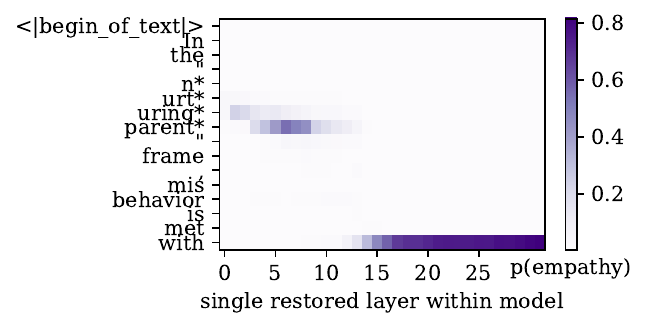} \\ 
\end{tabular}
\caption{  Impact of restoring hidden state (for a single layer and token) on Llama-3-8B-Instruct's prediction (`punishment' or `empathy') for the two prompts 
despite corrupting the subject tokens (SF/NP) in each prompt.}
\label{figure-rome}
\end{figure}

The results are presented in Fig. \ref{figure-rome}. 
The x-axis represents the layers of the model (from 0 to 31), 
the y-axis lists the tokens in the input prompt, 
and the expected correct output token is shown on the last row (right).
The color intensity represents the probability of the correct output token 
being restored when that specific hidden state is restored.
In line with our hypothesis and the literature, we find that the information about 
the SF/NP frames 
can be restored at two points:
the early layers (for the last subject token), 
and the later layers (for the last prompt token).

\subsubsection{Sparse Probing.}
Based on the above results, 
we can now focus on probing the hidden representation specifically at layer 17.\footnote{
Layer 17 is one of the first dark layers in the `late site' of Fig. \ref{figure-rome}.}

We use a binary logistic regression probe for our analysis  
due to its simplicity and ability to identifying salient dimensions 
without over-fitting with a complex probe.
To ensure our probe is really detecting the frames from the hidden representation,
and not memorizing our dataset, 
we further limit the number of 
dimensions of the hidden representation that are used 
with recursive feature elimination to five and one (making it sparse). 

We refine our dataset for the probing by preparing two binary classification tasks based on Table~\ref{table-zero-shot}: SF vs. control and NP vs. control.  
Further, we only keep SF texts that evoke the SF frame according to both humans and machines, 
and NP texts that evoke the NP frame according to both humans and machines, 
and make sure the control group evokes neither frame. 
To have balanced classes, we expand 
the number of SF and NP texts by asking GPT4 to generate approx. 20 
additional stories (final n$\approx$39 per group; with 20\% heldout).

The results for this experiment are shown in Table \ref{table-hiddendim}. 
What we find fascinating about these results is that we can tell the texts apart with an F1 score of around 80\%  with just 1 dimension out of the model's 4096 hidden dimensions. 

\begin{table*}[tb]
\caption{Statistics for a classifier probe that can identify texts that evoke the SF (or NP) frames from controls, using a limited number of hidden dimensions from the internal representation of  layer 17 for the last token. }
\label{table-hiddendim}
\begin{center}
\begin{small}
\begin{tabular}{lccc}
\hline  
\textsc{Frame} & \textsc{F1 score, 5 features} & \textsc{F1 score, 1 feature} & \textsc{Top Hidden Dim}   \\
\hline  
Strict Father & 0.93 (train) / 0.94 (test) & 0.78 / 0.81 & 133 \\ 
Nurturing Parent & 0.90 / 0.88 & 0.80 / 0.88  & 529 \\ 
\hline  
\end{tabular}
\end{small}
\end{center}
\end{table*}

The model may well be using shortcuts given the sample size--for instance, it might be detecting `authority' or `punishment' in the SF texts and `care' in the NP texts. When we look at the top five (tokenized) words in the SF texts, they are\footnote{This list excludes stop words and common words.}:  father (all), discipline (22 times) , children (15 times), figure (10 times), and son (10 times). The top five words in the NP texts are: children (20 times), parent (19 times), child (17 times), love (15 times), and care (14 times). But every token will still have numerous dimensions in the internal representation space, and the fact that we can detect one distinctive dimension suggests that the SF/NP elements enforce each other.


\section{Discussion \& Conclusion} 
\label{sec:discussion}

We found through our experiments and annotations that LLMs are generally fluent in a variety of socio-political frames. 
They can generate coherent original stories and or quote texts from other sources that evoke a frame---with all necessary sub-concepts of that frame.

It is perhaps not surprising, given the significance of frames in 
linguistics, psychology, and cognitive science, that machines that humans find
understandable are fluent in frames. 
However, this still has important ethical and societal implications.
Specifically, understanding how LLMs handle cognitive frames can have the positive consequence of demystifying the technology for a broader public. 
But it also suggests that the models may be able to
create persuasive misinformation.

We further offered some initial investigations into the mechanics of the frame generation. 
Notably, we found that information about the frame persists throughout the generation process,
specifically within the mid-layer hidden representations of the last prompt token.

This interdisciplinary study has been exploratory in nature, 
opening up areas for further research.
We found, for one, that LLMs are not equal in their abilities with regard
to cognitive frames, even at similar parameter sizes. 
This might be an area for further research, 
including testing more frames and model sizes,
and looking at the training data where possible
to better understand how LLMs learn abstract socio-political frames.

Regarding the mechanics, it would be interesting to investigate if the presence of a frame can be decreased or increased---in line with similar efforts in AI safety research---to perhaps remove undesirable frames from a discourse.

In conclusion, our study reveals that LLMs possess a general fluency in 
generating and recognizing complex socio-political frames, opening up new avenues 
for both NLP researchers and social scientists. 
While this capability demonstrates the advanced nature of these models, it also underscores the need for careful consideration of their societal impact. 

\begin{credits}
\subsubsection{\ackname} The authors are grateful to numerous colleagues who gave feedback on various stages of the research, including Jörg Pohle, Theresa Züger, and Freya Hewett. This study was funded by a grant from the German Federal Ministry of Science and Education (BMBF).  

\subsubsection{\discintname} The authors have no competing interests to declare that are
relevant to the content of this article.
\end{credits}

%
%
\bibliographystyle{splncs04}
\bibliography{references}

\appendix
\section{Additional Details on Prompts}

\subsection{Generation Prompts}
The following prompts were used with \texttt{temperature=0.7} to generate stories:

\begin{itemize}
    \item \sloppy\texttt{Please write a short original story which evokes/invokes the "XXX" frame (max one paragraph).}
    \item \texttt{Please pick a short passage from the Bible which evokes/invokes the "XXX" frame (max one paragraph).}
    \item \texttt{Please describe a short story from a scifi novel or movie which evokes/invokes the "XXX" frame (max one paragraph).}
\end{itemize}

Importantly: no additional context or instructions were provided in the prompts about the frame or the task.

\subsection{Zero-shot Prompts}
The following prompts were used with \texttt{temperature=0} to recognize frames:

\begin{itemize}
    \item \sloppy\texttt{Can you tell me which major cognitive frames are evoked by the following text? (Please keep your answer strictly short and name max 5 frames with no explanation)}
    \item \texttt{What percentage does the following text evoke the "XXX" frame? (Please give just the percentage with no additional words)}
 \end{itemize}
 
The following prompts were used to explore definitions and characteristics of the frames known to the (different) LLMs. 
The results helped
pick the contrasting auto-completion sentences used in 
Section \ref{section-hiddendim}.

\begin{itemize}
    \item \texttt{Please give a very short description of the "XXX" frame.}
    \item \texttt{Please describe key characteristics of the "XXX" frame.}
 \end{itemize}

\end{document}